\RequirePackage{fix-cm}
\documentclass[twocolumn]{svjour3}          
\smartqed  
\usepackage{graphicx}
\usepackage{amsmath} 
\usepackage{amssymb}  
\usepackage{hyperref}
\usepackage[table]{xcolor}
\usepackage{float}
\usepackage{setspace}
\usepackage{fancyhdr}
\usepackage{multirow}

\DeclareMathOperator{\EX}{\mathbb{E}}
\begin{document}

\title{Balancing reconstruction error and Kullback-Leibler divergence in Variational Autoencoders
}


\author{Andrea Asperti        \and
        Matteo Trentin 
}


\institute{A. Asperti \at
               University  of Bologna\\
Department of Informatics: Science and Engineering (DISI)\\
              \email{andrea.asperti@unibo.it}           
           \and
           M. Trentin \at
              University  of Bologna\\
Department of Informatics: Science and Engineering (DISI)
              \email{matteo.trentin@studio.unibo.it}     
}

\date{Received: date / Accepted: date}

\maketitle

\begin{abstract}
In the loss function of Variational Autoencoders
there is a well known tension between two components: 
the reconstruction loss, improving the quality of the resulting images, and
the Kullback-Leibler divergence, acting as a regularizer of the latent space. Correctly balancing
these two components is a delicate issue, easily resulting in poor generative
behaviours. In a recent work \cite{TwoStage}, a sensible improvement has been obtained by
allowing the network to {\em learn} the balancing factor during training, 
according to a suitable loss function.
In this article, we show that learning can be replaced by a simple deterministic
computation, helping to understand the underlying mechanism, and resulting in a faster and more 
accurate behaviour. On typical
datasets such as Cifar and Celeba, our technique sensibly outperforms all 
previous VAE architectures. 
\end{abstract}

\section{Introduction}
\label{sec:intro}
Generative models address the challenging task of capturing the 
probabilistic distribution of high-dimensional data, in order to gain insight in
their characteristic manifold, and ultimately paving the way to the 
possibility of synthesizing new data samples. 

The main frameworks of generative models that have been 
investigated so far are Generative Adversarial Networks (GAN) \cite{GAN} and Variational Autoencoders (VAE) (\cite{Kingma13,RezendeMW14}), both of which generated an enormous amount 
of works, addressing variants, theoretical investigations, or practical applications. 

The main feature of Variational  Autoencoders  is that they offer a strongly principled probabilistic approach to generative modeling. The key insight is the idea of addressing 
the problem of learning representations as a variational inference problem, coupling the generative model $P(X|z)$ for $X$ given the latent variable $z$, with an inference model $Q(z|X)$ synthesizing the latent representation of the given data.  

The loss function of VAEs is composed of two parts: one is just the log-likelihood of
the reconstruction, while the second one is a term aimed to enforce a {\em known} prior distribution $P(z)$ of the latent space - typically a spherical normal distribution.
Technically, this is achieved
by minimizing the Kullbach-Leibler distance between $Q(z|X)$ and the prior
distribution $P(z)$; as a side effect, this will also improve the similarity of the 
aggregate
inference distribution $Q(z) = \EX_X Q(z|Z)$ with the desired prior, that is our
final objective. 
\[ \hspace{.5cm} \underbrace{\EX_{z\sim Q(z|X)}log(P(X|z))}_{log-likelihood} - \lambda \cdot \underbrace{KL(Q(z|X)||P(z))}_{KL-divergence} \]
Loglikelihood and KL-divergence are typically balanced by a suitable $\lambda$-parameter (called $\beta$ in the terminology of $\beta$-VAE \cite{beta-vae17,understanding-beta-vae18}), since they have somewhat contrasting effects: the former will
try to improve the quality of the reconstruction, neglecting the shape of the latent 
space; on the other side, KL-divergence is normalizing and smoothing the latent 
space, possibly at the cost of some additional ``overlapping" between latent variables,
eventually resulting in a more noisy encoding \cite{brokenELBOW}. If not properly tuned, KL-divergence
can also easily induce a sub-optimal use of network capacity, where only
a limited number of latent variables are exploited for generation: this is the so called overpruning/variable-collapse/sparsity 
phenomenon \cite{BurdaGS15,overpruning17}. 

Tuning down $\lambda$ typically reduces the number of collapsed variables and improves the quality of {\em reconstructed} images. However, 
this may not result in a better quality of {\em generated samples}, since we
loose control on the shape of the latent space, that becomes harder to 
be exploited by a random generator. 

Several techniques have been considered for the correct calibration of $\gamma$, 
comprising an annealed optimization schedule \cite{Bowman15} or a policy 
enforcing minimum KL contribution from subsets of latent units \cite{autoregressive16}. 
Most of these schemes require hand-tuning and, quoting 
\cite{overpruning17}, they easily risk to ``take away the principled regularization scheme that is built into VAE.''

An interesting alternative that has been recently introduced in \cite{TwoStage} consists
in {\em learning} the correct value for the balancing parameter
during training, that also allows
its automatic calibration along the training process. The parameter is called $\gamma$, in this context, 
and it is considered as a normalizing factor for the reconstruction loss.

Measuring the trend of the loss function and of the learned lambda parameter during training, 
it becomes evident that the parameter is proportional to the reconstruction error, with the result that 
the relevance of the KL-component inside the whole loss function becomes independent
from the current error. 

Considering the shape of the loss function, it is easy to
give a theoretical justification for this behavior. As a consequence, 
there is no need for learning, that can be replaced by a simple deterministic
computation, eventually resulting in a faster and more accurate behaviour.

The structure of the article is the following.
In Section~\ref{sec:VAE}, we give a quick introduction to Variational Autoencoders, with particular emphasis on generative issues (Section~\ref{Sec:generation}).  In Section~\ref{Sec:balancing}, we discuss our approach to the problem of balancing
reconstruction error and Kullback-Leibler divergence in the VAE loss
function; this is obtained from a simple theoretical investigation of the loss function in \cite{TwoStage}, and essentially amounts to keeping a {\em constant} balance between
the two components along training. Experimental results are provided in Section~\ref{Sec:experimental}, relative to standard datasets such as CIFAR-10 (Section~\ref{Sec:cifar}) and CelebA (Section~\ref{Sec:celeba}): up to our knowledge, we get the best generative scores in terms of Frechet Inception Distance ever obtained by means of Variational Autoencoders. 
In Section~\ref{Sec:discussion}, we 
try to investigate the reasons why our technique seems to be more effective than previous approaches, by considering the evolution of
latent variables along training. Concluding remarks and ideas for future investigations are offered in Section~\ref{Sec:conclusions}.

\section{Variational Autoencoders}\label{sec:VAE}
In a generative setting, we are interested to express the probability of a data point $X$ through marginalization over a vector of latent variables:
\begin{equation}
\hspace{.5cm}P(X)  = \int P(X|z)P(z)dz \label{eq1} \approx \EX_{z\sim P(z)}P(X|z)
\end{equation}
For most values of $z$, $P(X|z)$ is likely to be close to zero, contributing in a 
negligible way in the estimation of $P(X)$, and hence making this kind of sampling 
in the latent space practically unfeasible.
The variational approach exploits sampling from an auxiliary ``inference" distribution $Q(z|X)$, hopefully producing values for $z$ more likely to effectively contribute to
the (re)generation of $X$. 
The relation between $P(X)$ and $\EX_{z\sim Q(z|X)}P(X|z)$ is given by the
following equation, where KL denotes the Kulback-Leibler divergence:
\begin{equation}
\hspace{.3cm}\begin{array}{l}
    log(P(X)) - KL(Q(z|X)||P(z|X)) = \\
    \hspace{.5cm}\EX_{z\sim Q(z|X)}log(P(X|z) - KL(Q(z|X)||P(z))\label{eq:elbo}
    \end{array}
\end{equation}
KL-divergence is always positive, so the term on
the right provides a lower bound to the loglikelihood $P(X)$, 
known as Evidence Lower Bound (ELBO).

If $Q(z|X)$ is a reasonable approximation of $P(z|X)$,  
the quantity $KL(Q(z)||P(z|X))$ is small; in this case the loglikelihood
$P(X)$ is close to the Evidence Lower Bound: the learning objective
of VAEs is the maximization of the ELBO.


In traditional implementations, we additionally assume that $Q(z|X)$ is 
normally distributed around an encoding
function $\mu_\theta(X)$, with variance $\sigma^2_\theta(X)$; similarly 
$P(X|z)$ is normally distributed around a decoder function $d_\theta(z)$.
The functions $\mu_\theta$, $\sigma^2_\theta$ and $d_\theta$ are approximated 
by deep neural networks. Knowing the variance of latent variables allows sampling
during training. 

Provided the model for the decoder function $d_\theta(z)$ is sufficiently expressive, the shape of the prior distribution $P(z)$ for latent variables can be arbitrary, and
for simplicity we may assumed it is a normal distribution $P(z) = G(0,1)$.
The term $KL(Q(z|X)||P(z)$ is hence the KL-divergence between two Gaussian distributions $G(\mu_\theta(X),\sigma^2_\theta(X))$ and $G(1,0)$ which can be 
computed in closed form:
\begin{equation}\label{eq:closed-form}
\hspace{.3cm}\begin{array}{l}
    KL(G(\mu_\theta(X),\sigma_\theta(X)),G(0,1)) = \\
    \hspace{1cm}\frac{1}{2}(\mu_\theta(X)^2 + \sigma^2_\theta(X)-log(\sigma^2_\theta(X)) -1)
\end{array}
\end{equation}
As for the term $\EX_{z\sim Q(z|X)}log(P(X|z)$, under the Gaussian assumption the logarithm of $P(X|z)$ is just the quadratic distance between $X$ and its reconstruction $d_\theta(z)$;
the $\lambda$ parameter balancing reconstruction error and KL-divergence can understood
in terms of the variance of this Gaussian (\cite{tutorial-VAE}).

The problem of integrating sampling 
with backpropagation, is solved by the well known reparametrization trick (\cite{VAE13,RezendeMW14}).

\subsection{Generation of new samples}\label{Sec:generation}
The whole point of VAEs is to force the generator to produce a marginal distribution\footnote{called by some authors {\em aggregate posterior distribution} \cite{AAE}.} $Q(z)  = \EX_{X}Q(z|X)$ close to the prior $P(z)$. If we average the
Kullback-Leibler regularizer $KL(Q(z|X)||P(z))$ on all input data, and expand 
KL-divergence in terms of entropy, we get:
\begin{equation}\label{eq:averaging}
  \hspace{.5cm}\begin{array}{l}
    \EX_{X}KL(Q(z|X)||P(z))\smallskip\\\smallskip
    = - \EX_{X} \mathcal{H}(Q(z|X)) + \EX_{X} \mathcal{H}(Q(z|X),P(z))\\\smallskip
    = - \EX_{X} \mathcal{H}(Q(z|X)) + \EX_{X} \EX_{z \sim Q(z|X)}log P(z)\\\smallskip
    = - \EX_{X} \mathcal{H}(Q(z|X)) + \EX_{z \sim Q(z)}log P(z)\\\smallskip
    = - \underbrace{\EX_{X} \mathcal{H}(Q(z|X))}_{\substack{\mbox{Avg. Entropy}\\\mbox{of } Q(z|X)}} +
          \underbrace{\mathcal{H}(Q(z),P(z))}_{\substack{\mbox{Cross-entropy }\\\mbox{of }Q(X) \mbox{ vs }P(z)}}
  \end{array}
  \end{equation}
  The cross-entropy between two distributions is minimal when they coincide, so we are pushing $Q(z)$ towards $P(z)$. At the same time, we try to augment the entropy of each $Q(z|X)$; under the assumption that $Q(z|X)$ is Gaussian, this amounts to 
  enlarge the variance, further improving the coverage of the 
  latent space, essential for generative sampling (at the cost
  of more overlapping, and hence more confusion between the 
  encoding of different datapoints). 

Since our prior distribution is a Gaussian, we expect $Q(z)$ to be normally distributed too,
so the mean $\mu$ should be 0 and the variance $\sigma^2$ should be 1.
If $Q(z|X) = N(\mu(X),\sigma^2(X))$, we may look at $Q(z) = \EX_X Q(z|X)$ as a Gaussian Mixture Model (GMM). 
Then, we expect
\begin{equation}
\hspace{3cm}\EX_X \mu(X) = 0
\end{equation}
and especially, assuming the previous equation (see \cite{aboutVAE} for details), 
 \begin{equation}
\hspace{2cm}\sigma^2 = \EX_X \mu(X)^2 +  \EX_X \sigma^2(X)  = 1
\end{equation}
This rule, that we call {\em variance law}, provides a simple sanity check to test
if the regularization effect of the KL-divergence is properly working. 

The fact that the two first moments of the marginal inference distribution 
are 0 and 1, does not imply that it should look like a Normal. The possible
mismatching between $Q(z)$ and the expected prior $P(z)$ is indeed a problematic 
aspect of VAEs that, 
as observed in 
several works \cite{ELBOsurgery,rosca2018distribution,aboutVAE} could compromise
the whole generative framework. To fix this, some
works extend the VAE objective by encouraging the aggregated posterior to match $P(z)$
\cite{WAE} or by exploiting more complex priors \cite{autoregressive,Vamp,resampledPriors}.

In \cite{TwoStage} (that is the current state of the art), a
second VAE is trained to learn an accurate approximation of $Q(z)$; 
samples from a Normal distribution are first used to generate samples
of $Q(z)$, that are then fed to the actual generator of data points. Similarly, 
in \cite{deterministic}, the authors try to give an ex-post estimation of $Q(z)$, e.g. imposing 
a distribution with a sufficient complexity (they consider a combination of 
10 Gaussians, reflecting the ten categories of MNIST and Cifar10).

\section{The balancing problem}\label{Sec:balancing}
As we already observed, the problem of correctly balancing reconstruction error
and KL-divergence in the loss function has been the object of several investigations.
Most of the approaches were based on empirical evaluation, and often required
manual hand-tuning of the relevant parameters. A more theoretical approach has been 
recently pursued in \cite{TwoStage} 

The generative loss (GL), to be summed with the KL-divergence, is defined by the following
expression (directly borrowed from the public code\footnote{\url{https://github.com/daib13/TwoStageVAE}.}):
\begin{equation}\label{eq:GL}
\hspace{2cm}GL = \frac{\mbox{mse}}{2\gamma^2} + log \gamma + \frac{log 2\pi}{2}\end{equation}
where mse is the mean square error on the minibatch under consideration
and $\gamma$ is a parameter of the model, learned during training.
The previous loss is derived in \cite{TwoStage} by a complex analysis of
the VAE objective function behavior, assuming the decoder has a gaussian error with
variance $\gamma^2$, and investigating the case of arbitrarily small but explicitly 
nonzero values of $\gamma^2$.

Since $\gamma$ has no additional constraints, we can explicitly minimize it
in equation \ref{eq:GL}. The derivative $GL'$ of $GL$ is
\begin{equation}\label{eq:GL'}
\hspace{2.5cm}GL' = -\frac{\mbox{mse}}{\gamma^3} + \frac{1}{\gamma}\end{equation}
having a zero for $\gamma^2 = \mbox{mse}$, corresponding to a minimum for equation \ref{eq:GL}. 

This suggests a very simple deterministic policy for {\em computing} $\gamma$ instead of
{\em learning} it: just use the current estimation of the mean square error. This 
can be easily computed as a discounted combination of the mse relative to the current 
minibatch with the previous approximation: in our implementation, we just take
the minimum between these two values, in order to have a monotically decreasing 
value for $\gamma$ (we work with minibatches of size 100, that is sufficiently 
large to provide a reasonable approximation of the real mse). 
Updating is done at every minibatch of samples. 

Compared with the original approach in \cite{TwoStage}, the resulting technique is
both faster and more accurate. 

An additional contribution of our approach is to bring some light on the effect of
the balancing technique in \cite{TwoStage}. Neglecting constant addends, that 
have no role in the loss
function, the total loss function for the VAE is simply:
\begin{equation}\label{eq:totalloss}
\hspace{3cm}GL = \frac{\mbox{mse}}{2\gamma^2} + KL\end{equation}

So, computing gamma according to the previous estimation of $\mbox{mse}$ has essentially
the effect of keeping a {\em constant} balance between reconstruction error and KL-divergence
during the whole training: as mse is decreasing, we normalize it in order to prevent a prevalence 
of the KL-component, that would forbid further improvements of the quality of
reconstructions.

\section{Empirical evaluation}\label{Sec:experimental}
\begin{figure}
\hspace{-.3cm}\begin{tabular}{cc}
\includegraphics[width=.23\textwidth]{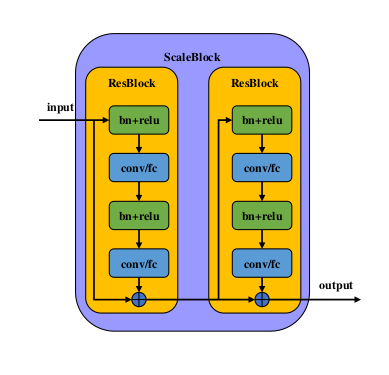}&
\hspace{-.2cm}\includegraphics[width=.23\textwidth]{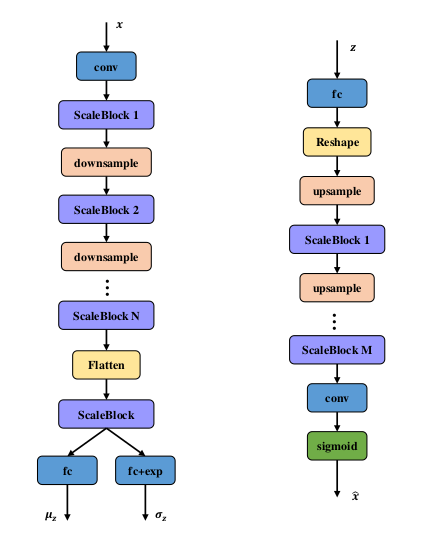}\\
(A) Scale block & (B) Encoder \;\;(C) decoder
\end{tabular}
\caption{\label{fig:resnet}"Resnet'' architecture. 
(A) Scale block: a sequence of residual blocks. We mostly worked with a single residual block; two or more blocks makes the architecture sensibly heavier and slower to train, with no remarkable improvement (B) Encoder: the input is first
transformed by a convolutional layer into and then passed to a chain of Scale blocks;
after each Scale block, input is downsampled with a a convolutional layer with stride
$2$ channels are doubled. After $N$ Scale blocks, the feature map is flattened to a vector.
and then fed to another Scale Block composed by fully connected layers of
dimension 512. The output of this Scale Block is used to produce mean and variances 
of the $k$ latent variables. Following \cite{TwoStage}, $N=3$ and $k=64$ for CIFAR-10.
For CelebA, we tested many different configurations. 
(C) Decoder: the latent representation $z$ is
first passed through a fully connected layer, reshaped to 2D, and then passed through
a sequence of deconvolutions halving the number of channels at the same.}
\end{figure}

We compared our proposed Two Stage VAE with computed $\gamma$ against the
original model with learned $\gamma$ using the same network architectures.
In particular, we worked with many different variants of the so called ResNet version, schematically described in 
Figure~\ref{fig:resnet} (pictures are borrowed from \cite{TwoStage}).
In all our experiments, we used a batch size of 100, and adopted Adam with default TensorFlow's hyperparameters as optimizer. Other hyperparameters, as well as 
additional architectural details will be described below, where we discuss the cases 
of Cifar and CelebA separately. 

In general, in all our experiments, we observed a high sensibility of Fid scores 
to the learning rate, and to the deployment of auxiliary regularization techniques. 
As we shall discuss in Section~\ref{Sec:discussion}, 
modifying these training configurations may easily result in a different number of 
inactive\footnote{for the purposes of this work, we consider 
a variable {\em inactive} when $\EX_X \sigma^2(X) \ge .8$} latent variables at the end of training. 
Having both too few or too many active variables
may eventually compromise generative sampling, for opposite reasons: few active variables
usually compromise reconstruction quality, but an excessive number of active variables 
makes controlling the shape of the latent space sensbibly harder.

The code is available on \href{https://github.com/asperti/BalancingVAE.git}{GitHub}\footnote{\url{https://github.com/asperti/BalancingVAE.git}}. Checkpoints for Cifar10 and CelebA are available at the \href{http://www.cs.unibo.it/~asperti/balancingVAE}{project's page}\footnote{\url{http://www.cs.unibo.it/~asperti/balancingVAE}}.


\subsection{Cifar10}\label{Sec:cifar}
\begin{figure}[h!]
\centering{\includegraphics[width=.88\columnwidth]{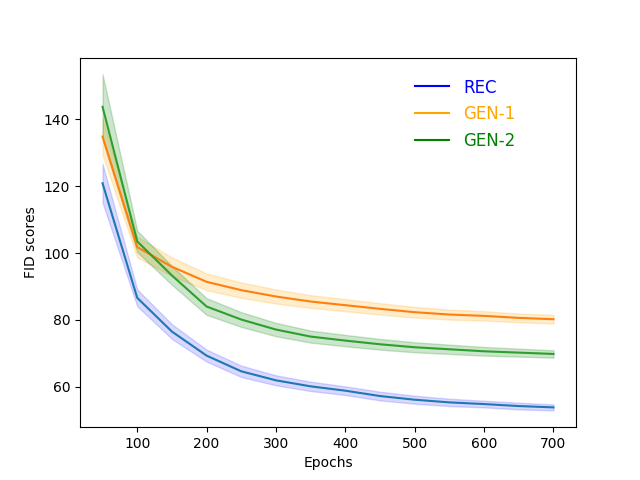}}
\caption{\label{fig:evolution}
Evolution during 700 epochs of training on the CIFAR-10 
dataset of the FID scores for reconstructed images (blue),
first-stage generated images (orange), and second-stage generated images. The number of
epochs refer to the first VAE, and it is doubled for the second VAE. The filled
region around the line corresponds to the standard deviation from the expected value.
Mean and variances have been estimated over 10 different trainings.}
\end{figure}

For Cifar10, we got relatively good results with the basic ResNet architecture
with 3 Scale Blocks, a single Resblock for every Scaleblock, and 64 latent
variables. We trained our model for 700 epochs on the first VAE and 1400 epochs 
on the second VAE; the initial learning rate was $0.0001$, halving it every $200$ epochs on the first VAE and every $100$ epochs on the
second VAE. Details about the evolution
of reconstruction and generative error during training are provided in 
Figure~\ref{fig:evolution} and  Table~\ref{tab:evolution}. The data refer to
ten different but ``uniform" trainings ending with the same number of active latent variables, 
(17 in this case). Few pathological trainings resulting in less or higher sparsity 
(and worse FID scores) have been removed from the statistic. 

\begin{table*}[ht!]
\caption{\label{tab:evolution}
Evolution during training on the CIFAR-10 
dataset of several different metrics}
    \begin{center}$
    \begin{array}{|r|c|c|c|c|c|}\hline
         \mbox{epochs} & \mbox{REC} & \mbox{GEN-1} & \mbox{GEN-2} 
           & \mbox{mse} & \mbox{variance law} \\\hline
         50 & 120.9 \pm 5.8 & 134.9 \pm 5.9 & 143.8 \pm 9.7
         & .0089 \pm .0002 & 1.012 \pm .012 \\\hline
         100 & 86.6 \pm 2.5 & 101.8 \pm 3.1 & 103.5 \pm 3.2 
         &  .0071 \pm .0001 & .969 \pm .010 \\\hline
         150 & 76.5 \pm 2.3 & 95.9 \pm 2.8 & 93.3 \pm 2.8
         & .0062 \pm .0001 & .939 \pm .011 \\\hline
         200 & 69.3 \pm 2.1 & 91.4 \pm 2.5 & 84.0 \pm 2.5
         & .0056 \pm .0001 & .918 \pm .012 \\\hline
         250 & 64.6 \pm 2.0 & 88.9 \pm 2.3 & 80.1 \pm 2.2
         & .0052 \pm .0001 & .922 \pm .013 \\\hline
         300 & 61.9 \pm 1.8 & 87.0 \pm 2.1 & 77.1 \pm 2.0
         & .0049 \pm .0001 & .925 \pm .012 \\\hline
         350 & 60.1 \pm 1.7 & 85.5 \pm 1.9 & 75.0 \pm 1.8
         & .0047 \pm .0001 & .928 \pm .010 \\\hline
         400 & 58.8 \pm 1.5 & 84.4 \pm  1.8 & 73.8 \pm 1.7
         & .0046 \pm .0001 & .930 \pm .008 \\\hline
         450 & 57.2 \pm 1.3 &  83.3 \pm 1.7 & 72.7 \pm 1.6
         & .0045 \pm .0001 & .937 \pm .007 \\\hline
         500 & 56.1 \pm 1.2 & 82.3 \pm 1.6 & 71.8 \pm  1.5 
         & .0044 \pm .0001 & .942 \pm .007 \\\hline
         550 & 55.3 \pm 1.1 & 81.7 \pm 1.5 & 71.2 \pm 1.4
         & .0043 \pm .0001 & .946 \pm .007 \\\hline
         600 & 54.8 \pm 1.1 & 81.2 \pm 1.4 & 70.6 \pm 1.3
         & .0043 \pm .0001 & .950 \pm .006 \\\hline
         650 & 54.2 \pm 1.0 & 80.7 \pm 1.4 & 70.2 \pm 1.2
         & .0042 \pm .0001 & .954 \pm .005 \\\hline
         700 & 53.8 \pm 0.9 & 80.2 \pm 1.3 & 69.8 \pm 1.1
         & .0041 \pm .0001 & .957 \pm .005 \\\hline
    \end{array}
    $\end{center}
\end{table*}

\begin{table*}[ht!]
\caption{\label{tab:results}CIFAR-10: summary of results}
\begin{center}
\begin{tabular}{|c|c|c|c|c|}\hline
      model   & epochs & REC & GEN-1 & GEN-2\\\hline
RAE-l2 \cite{deterministic} (128 vars)& 100 & $32.24 \pm ? $ & $80.8 \pm ? $ & $74.2 \pm ? $ \\\hline
2S-VAE, learned $\gamma$ \cite{TwoStage} &1000 & & $76.7 \pm 0.8$ & $72.9 \pm 0.9$\\
2S-VAE, learned $\gamma$, replicated & 1000 & $54.1 \pm 0.9$ & $76.8 \pm 1.2$ & $73.1 \pm 1.2$ \\\hline
2S-VAE, computed $\gamma$ & 700 & $53.8 \pm 0.9$ & $80.2 \pm 1.3$ & ${\bf 69.8} \pm 1.1$\\\hline

\end{tabular}
\end{center}
\end{table*}
In Table~\ref{tab:results}), we compare our approach with the original version with learned $\gamma$\cite{TwoStage}. Since some people had problems in replicating the results in
\cite{TwoStage}
(see the discussion on \href{https://openreview.net/forum?id=B1e0X3C9tQ}{OpenReview}\footnote{\url{https://openreview.net/forum?id=B1e0X3C9tQ}}), we repeated the experiment (also in 
order to compute the reconstruction FID). Using
the learning configuration suggested by the authors, namely 1000 epochs for the first VAE, 
2000 epochs for the second one, 
initial learning rate equal to 0.0001, halved every 300 and 600 
epochs for the two stages, respectively, we obtained results essentially in line with those declared in \cite{TwoStage}.

\begin{table*}[ht]
\caption{\label{tab:celeba1}CelebA: metrics for different models}
\vspace{-.2cm}
    \[
    \begin{tabular}{|c|c||c|c|c|c|c||c|c|c|c|c|}\hline
               & & \multicolumn{5}{c|}{learned $\gamma$} & \multicolumn{5}{c|}{computed $\gamma$}\\\hline
         Model & epochs & REC & GEN-1 & GEN-2 & mse & var.law & REC & GEN-1 & GEN-2 & mse & var.law \\\hline
         1 & 40/120 & 53.8 & 66.0 & 59.3 & .0059 & 1.024 & 45.8 & 56.9 & 57.1 & .0056 & 0.805 \\\hline
         1 & 80/210 & 54.3  & 65.9 & 59.8 & .0049 & 0.803 & 46.0 & 61.1 & 58.1 & .0047 & 0.688 \\\hline
         1 & 120/300 & 54.9  & 66.5 & 60.4 & .0044  & 0.775 & 48.1 & 63.8 & 59.9 & .0043 & 0.687 \\\hline\hline
         2 & 40/120 & 48.5 & 58.2 & 54.5 & .0059 &  0.985 & 41.5 & 58.7 & 53.7 & .0058 & 1.024 \\\hline
         2 & 80/210 & 48.8 & 60.7 & 55.5 & .0048 & 0.889 & 42.0 & 58.8 & 55.1 & .0048 & 0.877 \\\hline
         2 & 120/300 & 49.1 & 62.8 & 56.9 & .0043  & 0.880 & 43.0 & 60.2 & 56.2 & .0043 & 0.863 \\\hline\hline
         3 & 40/120 & 59.4  & 74.3 & 63.4 & .0050  & 0.893 & 56.3 & 74.0 & 63.9 & .0049 & 0.637 \\\hline
         3 & 80/210 & 55.6  & 72.6 & 62.2 & .0039 & 0.840 & 54.4 & 72.3 & 61.8 & .0038 & 0.621 \\\hline
         3 & 120/300 & 55.2 & 72.0 & 62.1 & .0037  & 0.785 & 54.4 & 71.3 & 62.0 & .0036 & 0.744 \\\hline\hline
         4 & 40/120 &  52.8 & 68.2 & 60.4 & .0072 & 0.789 & 48.0 & 65.0 & 57.7 & .0072 & 0.742 \\\hline
         4 & 80/210 &  49.4 & 67.9 & 58.5 & .0060 & 0.822 & 45.3 & 65.0 & 53.4 & .0059 & 0.785 \\\hline
         4 & 120/300 & 49.1 & 68.0 & 58.4 & .0053  & 0.844 & 44.4 & 65.4 & 54.0 &.0053 & 0.804 \\\hline\hline
    \end{tabular}
    \]
\end{table*}

For the sake of completeness, 
we also compare with the FID scores for the recent RAE-l2 model \cite{deterministic}
(variance was not provided by authors). In this case, 
the comparison is purely indicative, since in \cite{deterministic} they work, in the CIFAR-10 case, with
a latent space of dimension 128. This also explains their
particularly good reconstruction error, and the few training epochs.

\subsection{CelebA}\label{Sec:celeba}
In the case of CelebA, we had more trouble in replicating the results 
of \cite{TwoStage}, although we were working with their own code. As we shall see, this was partly due to a mistake on our side, that pushed us to an extensive investigation of different architectures.

In Table~\ref{tab:celeba1} we summarize {\em some} of the results we obtained, over a large variety of different network configurations. 
The metrics given in the table refer to the following models:

\begin{itemize}
    \item Model 1: This is our base model, with 4 scale blocks in the first stage, 64 latent variables, and dense layers with inner dimension 4096 in the second stage. 
    \item Model 2: As Model 1 with l2 regularizer added in upsampling and scale layers in the decoder.
    \item Model 3: Two resblocks for every scale block, l2 regularizer added in downsampling layers in the encoder.
    \item Model 4: As Model 1 with 128 latent variables, and 3 scale blocks.
\end{itemize}
All models have been trained with Adam, with an initial learning rate of
0.0001, halved every 48 epochs in the first stage and every 120 epochs in the second stage.


According to the results in Table~\ref{tab:celeba1}, we can do a few noteworthy observations:
\begin{enumerate}
    \item for a given model, the technique computing $\gamma$ systematically outperforms the version learning it, both in reconstruction and generation on both stages;
    \item after the first 40 epochs, FID scores (comprising reconstruction FID) do not seem to improve any further, and can even get worse, in spite of the fact that the mean square error keep decreasing; this is in contrast with
    the intuitive idea that FID REC score should be proportional to mse;
    \item the variance law is far from one, that seems to suggest Kl is too weak, in this case; this justifies the mediocre generative scores of the first stage, and the sensible improvement obtained with the second stage;
    \item l2-regularization, as advocated in \cite{deterministic}, seems indeed to have some beneficial effect. 
\end{enumerate}

\begin{figure}
    \centering
    \includegraphics[width=.9\columnwidth]{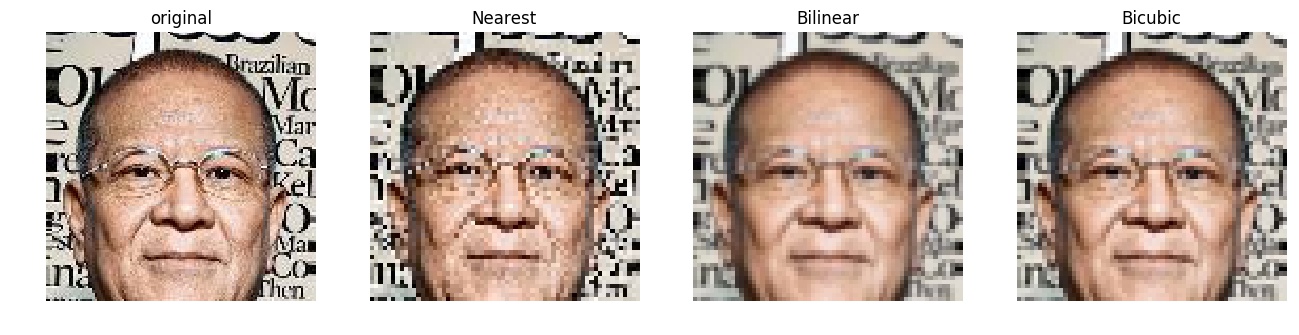}\\
    \includegraphics[width=.9\columnwidth]{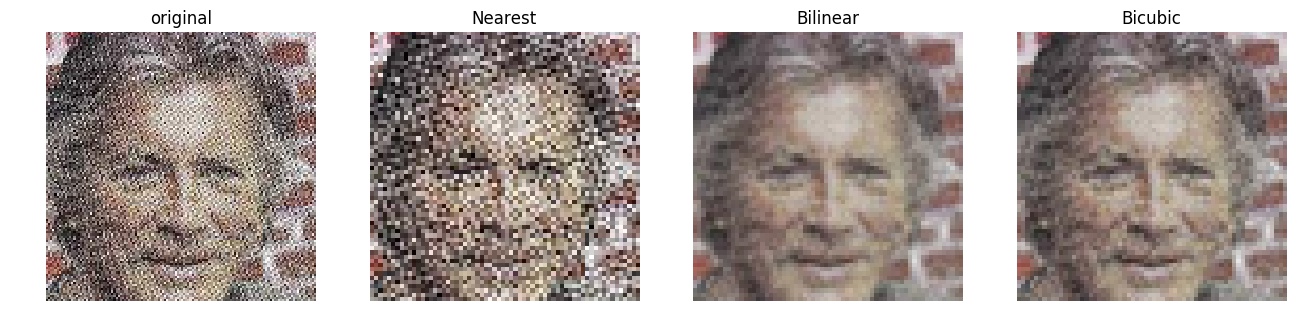}\\
    \includegraphics[width=.9\columnwidth]{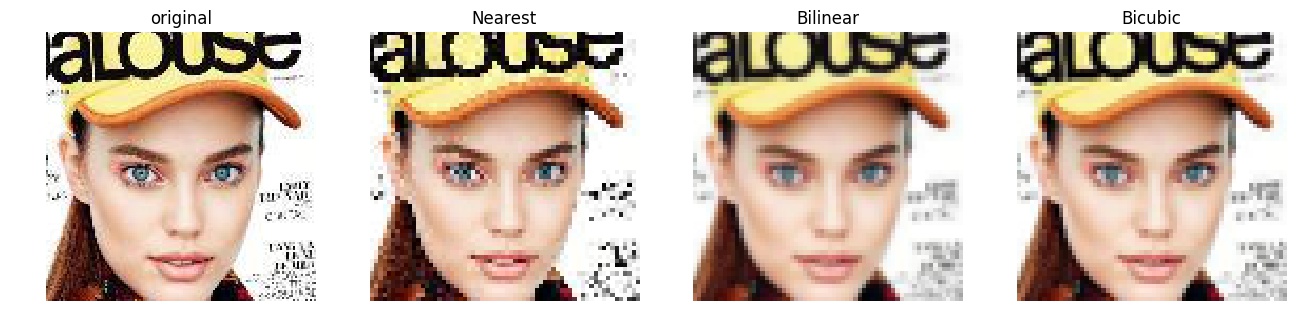}
    \caption{Effect of resizing mode on a few CelebA samples. Nearest Neighbours produces bad staircase effects; bilinear, that is the common choice, is particularly smooth, suiting well  to VAEs; bicubic is sligtly sharper.}
    \label{fig:resizing}
\end{figure}

We spent quite a lot of time trying to figure out the reasons of the discrepancy between our observations, and the results claimed in \cite{TwoStage}. 
Inspecting the elements of the dataset with worse reconstruction errors, we remarked a particularly bad quality of some of the images, resulting from the resizing of the face crop of dimension 128x128 to the canonical dimension 64x64 expected from the neural network. The resizing function used in the source code of \cite{TwoStage} available at was the deprecated \verb+imresize+ function of the scipy library\footnote{scipy imresize: \href{https://docs.scipy.org/doc/scipy-1.2.1/reference/generated/scipy.misc.imresize.html}{\url{https://docs.scipy.org/doc/scipy-1.2.1/reference/generated/scipy.misc.imresize.html}}}. Following the suggestion in the documentation, we replaced the call to imresize with a call to PILLOW:\\\smallskip
\verb+numpy.array(Image.fromarray(arr).resize())+\\\smallskip
Unfortunately, and surprisingly, the default resizing mode of PILLOW is Nearest Neighbours that, as described in Figure~\ref{fig:resizing}, introduces annoying jaggies that sensibly deteriorate the quality of images.
This probably also explains the anomalous behaviour of FID REC with respect to mean squared error. The Variational Autoencoder fails to reconstruct
images with high frequency jaggies, while keep improving on smoother images. This can be experimentally confirmed by the fact that while the minimum
mse keeps decreasing during training, the maximum, after a while, stabilizes. So, in spite of the fact that the average mse decreases, the overall distribution of reconstructed images may remain far from the distribution of real images, and possibly get even more more distant.

\begin{table*}[ht]
\caption{\label{tab:CelebAresults}CelebA: summary of results}
\begin{center}
\begin{tabular}{|c|c|c|c|c|}\hline
model        & epochs      & REC & GEN-1 & GEN-2\\\hline
RAE-SN \cite{deterministic} & 70 & $36.0 \pm ? $ & $44.7 \pm ? $ & $40.9 \pm ? $ \\\hline
2S-VAE, learned $\gamma$ \cite{TwoStage} & 120 &     & $60.5 \pm 0.6$ & $44.4 \pm 0.7$ \\\hline
2S-VAE, computed $\gamma$ & \multirow{2}{*}{70} & \multirow{2}{*}{${\bf 33.9} \pm 0.8$} & \multirow{2}{*}{${\bf 43.6} \pm 1.3$} & $42.7 \pm 1.0$\\
with latent space norm. & & & & ${\bf 38.6} \pm 1.0$ \\\hline

\end{tabular}
\end{center}
\end{table*}


Resizing images with the traditional bilinear interpolation produces a substantial improvement, but not sufficient to obtain the expected generative scores. 

Another essential component is again the balance between reconstruction error and KL-divergence. As observed above, in the case of
CelebA the KL-divergence seems too weak, as clearly testified by the moments of latent variables expressed by the variance law. 
As a matter of fact, in the loss function of \cite{TwoStage}, both mse and KL-divergence are computed as {\em reduced sums}, respectively over pixels and latent variables. Now, passing from CIFAR-10 to Celeba, we multiplied the number of pixels by four, passing from 32x32 to 64x64, but kept a constant number of latent variables. So, in order to keep the same balance we used for CIFAR-10, we should multiply the KL-divergence by a factor 4.

Finally, learning seems to proceed quite fast in the case of CelebA, that suggests to work with a lower initial learning rate: 0.00005. 
We also kept l2 regularization on downsampling and upsampling layers. 

With these simple expedients, we were already able to improve on generative scores in \cite{TwoStage}, 
(see Table~\ref{tab:CelebAresults}), but not with respect to \cite{deterministic}.

\begin{figure}[h!]
\includegraphics[width=\columnwidth]{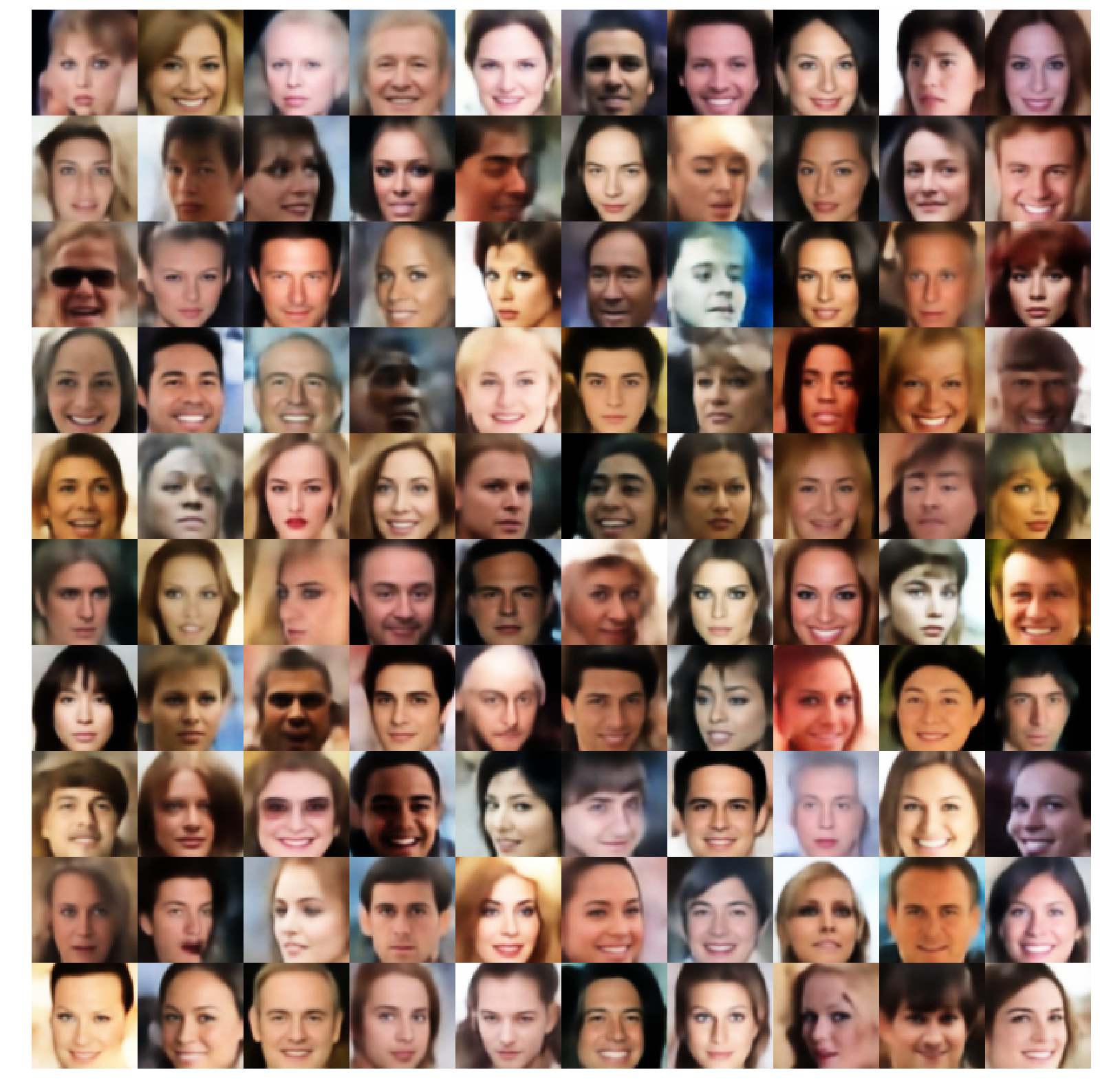}
\caption{\label{fig:celeba-gen}Examples of generated faces. The resulting images do not show the
blurred appearance so typical of variational approaches, sensibly improving their perceptive quality.}
\end{figure}

Analyzing the moments of the distribution of latent variables generated during the second
stage, we observed that the actual variance was sensibly below the expected unitary variance
(around .85). The simplest solution consists in normalizing the generated latent variables,
to meet the expected variance (this point is a bit outside the scope of this contribution,
and will be better investigated in a forthcoming article). 

This final precaution caused a sudden burst in the FID score for generated images, 
permitting to obtain, to the best of our knowledge, the best generative scores ever produced for CelebA with a variational approach.

In Figure~\ref{fig:celeba-gen} we provide examples of randomly generated faces. Note
the particularly sharp quality of the images, so unusual for variational approaches.

\section{Discussion}\label{Sec:discussion}
The reason why the balancing policy between reconstruction error and KL-regularization
addressed in \cite{TwoStage} and revisited in this article is so effective seems to
rely on its laziness in the choice of the latent representation. 

\begin{figure}[ht]
\includegraphics[width=.95\columnwidth]{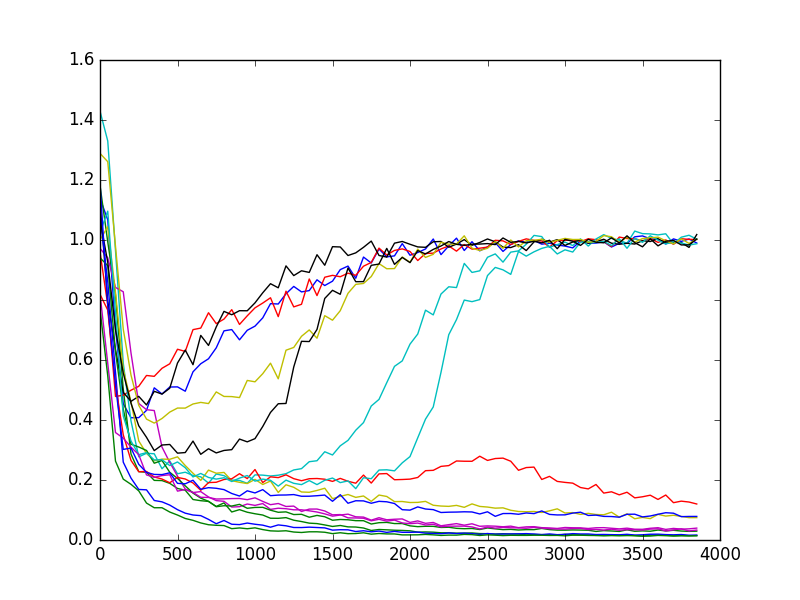}
\caption{\label{typical}Typical evolution of the mean variance $\EX_{X}\sigma^2_z(X)$ of latent variables during training
in a Variational Autoencoder.
Relevant variables have a variance close to $0$, while inactive variables have a
variance going to $1$. The picture was borrowed from \cite{sparsity} and is relative
to the first epoch of training for a dense VAE over the MNIST data set.}
\end{figure}

A Variational Autoencoder computes, for each latent variable $z$ and each sample $X$,
an expected value $\mu_z(X)$ and a variance $\sigma^2_z(X)$ around it. During training,
the variance $\sigma^2_z(X)$ usually drops very fast to values close to $0$, reflecting
the fact that the network is highly confident in its choice of $\mu_z(X)$. The
KL-component in the loss function can be understood as a mechanism aimed to reduce
this confidence, by forcing a not negligible variance. By effect of the KL-regularization,
some latent variables may be even neglected by the VAE, inducing {\em sparsity} in
the resulting encoding \cite{sparsity}. The ``collapsed" variables have, for any $X$, a
value of $\mu_z(X)$ close to $0$ and a mean variance $\sigma^2_z(X)$ close $1$.
So, typically, 
at a relatively early stage of training, the mean variance $\EX_{X}\sigma^2_z(X)$ of each
latent variable $z$ gets either close to $0$, if the variable is exploited, of close
to $1$ if the variable is neglected (see Figure~\ref{typical}). 

Traditional balancing policies addressed in the literature start with a low value
for the KL-regularization, increasing it during training. The general idea 
is to start privileging the quality of reconstruction, and then try to induce a better
coverage of the latent space. Unfortunately, this reshaping ex post of the latent
space looks hard to achieve, in practice.

The balancing property discussed in this article does the opposite: it starts attributing 
a relatively high importance to KL-divergence, to balance the high initial reconstruction error, progressively reducing its relevance in a way proportional to the improvement of the reconstruction.
In this way, the relative importance between the two components of the loss function
remains constant during training. 

The practical effect is that latent variables are kept for a long time in a sort of limbo
from which, one at a time, they are retrieved and put to work by the autoencoder, 
as soon as it realizes how they can contribute to the reconstruction.

The previous behaviour is evident by looking
at the evolution of the mean variance $\EX_{X}\sigma^2_z(X)$ of latent variables during training (not to be confused with the variance of
the mean values $\mu_z(X)$, that according to the variance law
should approximately be the complement to $1$ of the former).

\begin{figure}[ht]
\includegraphics[width=.95\columnwidth]{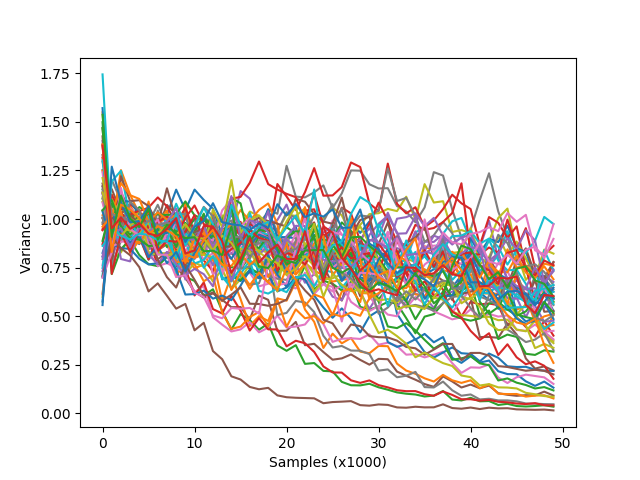}
\caption{\label{first_epoch}Evolution of the mean variance of the 64 latent variables
during the first epoch of training on Cifar10. Due to the "lazy" balancing technique, 
even after a full epoch, the destiny of most latent variables is still uncertain: they 
could collapse or be exploited for reconstruction.}
\end{figure}

In Figure~\ref{first_epoch} 
we see the evolution of the variance of the 64 latent
variables during the first epoch of training on the Cifar10 data set: even after a
full epoch, the ``status" of most latent variables is still uncertain.

\begin{figure}[ht]
\includegraphics[width=.95\columnwidth]{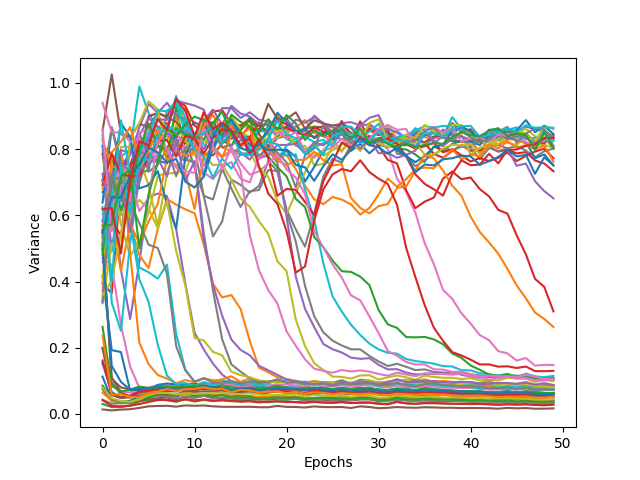}
\caption{\label{50_epochs}Evolution of the mean variance of the 64 latent variables
First 50 epochs of training on Cifar10. One by one, latent variables are retrieved
from the limbo (variance around 0.8) , and put to work by the autoencoder.}
\end{figure}

During the next 50 epochs, in a very slow process, some of the ``dormient'' latent variables
are woken up by the autoencoder, causing their mean variance to move towards 0:
see Figure~\ref{50_epochs}.

With the progress of training, less and less variables change their status, until
the process finally stabilizes. 

It would be nice to think, as hinted to in \cite{TwoStage},
that the number of active latent variables at the end of training corresponds to
the {\em actual dimensionality of the data manifold}. Unfortunately, this number still 
depends on too many external factors to justify such a claim. For instance, a mere
modification of the learning rate is sensibly affecting the sparsity of the resulting
latent space, as shown in Table~\ref{tab:inactive} where we compare, for different initial learning rates (l.r.), the final number of inactive variables, FID scores, 
and mean square error.

\begin{table}[h]\caption{Effect of the learning rate on sparsity and different metrics. 
A high learning rate reduces sparsity and improves on reconstruction. However, this does
not result in a better generative score. With a low rate, too many variables remains inactive.}
    \centering
    \begin{tabular}{|c|c|c|c|c|c|}\hline
         l.r. &  inact. & REC & GEN-1 & GEN-2 & mse  \\\hline
         .00020 & 13 & 53.0 & 80.6 & 74.5 & .0039 \\\hline
         .00015 & 15 & 53.3 & 79.9 & 71.8 & .0040 \\\hline
         .00010 & 17 & 53.8 & 80.2 & 68.8 & .0041 \\\hline
         .00005 & 19 & 58.2 & 83.2 & 75.8 & .0047 \\\hline
    \end{tabular}
    \label{tab:inactive}
\end{table}
Specifically, a high learning rate appears to be in conflict with 
the lazy way we would like latent variables to be chosen for activation; this typically 
results in less sparsity, that is not always beneficial for
generative purposes. 
The annoying point is that with respect to the dimensionality
of the latent space with the best generative FID, activating more variables can result in a lower reconstruction error, that should not be the case if we correctly
identified the datafold dimensionality. 

So, while the balancing strategy discussed
in this article (similarly to the one in \cite{TwoStage}) is eventually beneficial, 
still could take advantage of some tuning.

\subsection{Conclusions}\label{Sec:conclusions}
In this article, we stressed the importance of keeping a  constant balance between reconstruction error and Kullback-Leibler divergence during training of Variational Autoencoders. We did so by normalizing the reconstruction
error by an estimation of its current value, derived from
minibatches. We developed the technique by an
investigation of the loss function used in \cite{TwoStage},
where the balancing parameter was instead learned during
training. Our technique seems to outperform all previous
Variational Approaches, permitting us to obtain unprecedented
FID scores for traditional datasets such as CIFAR-10 and 
CelebA.

In spite of its relevance, the politics of keeping a constant balance does not seem to entirely solve the balancing issue, 
that still seems to depend from many additional factors, such as the network
architecture, the complexity and resolution of the dataset, or from
training parameters, such as the learning rate. 

Also, the regularization effect of the KL-component must be better understood, since it frequently fails to induce the expected 
distribution of latent variables, possibly requiring and justifying ex-post adjustments. 
\bigskip

\noindent
{\bf Credits}: All innovative ideas and results contained in this article are to be credited to
the first author. The second author mostly contributed on the experimental side.
\bigskip

\noindent
{\bf Conflict of Interest}: The authors declare that they have no conflict of interest.

\bibliographystyle{plain}
\bibliography{machine.bib,variational.bib}

\end{document}